\documentclass[letterpaper]{article}

\usepackage{microtype}
\usepackage{graphicx}
\usepackage{subcaption}
\usepackage{booktabs} 
\usepackage{kotex}

\usepackage{hyperref}

\usepackage[accepted]{icml2026}
\makeatletter
\renewcommand{\ICML@appearing}{%
\textit{Accepted at the SPIGM Workshop at the 43rd International Conference on Machine Learning (ICML 2026)},
Seoul, South Korea, 2026.}
\makeatother
\usepackage{amsmath}
\usepackage{amssymb}
\usepackage{mathtools}
\usepackage{amsthm}

\usepackage[capitalize,noabbrev]{cleveref}

\theoremstyle{plain}
\newtheorem{theorem}{Theorem}[section]

\theoremstyle{definition}

\newtheorem{example}[theorem]{Example}
\theoremstyle{remark}

\usepackage[disable,textsize=tiny]{todonotes}

\icmltitlerunning{CFG-OEC: Classifier Free Guidance with Orthogonal Error Correction}

\begin{document}

\twocolumn[
  \icmltitle{CFG-OEC: Classifier Free Guidance with Orthogonal Error Correction} 



  \icmlsetsymbol{equal}{*}

  \begin{icmlauthorlist}
    \icmlauthor{Nakgyu Yang}{equal,KAIST}
    \icmlauthor{Yechan Lee}{equal,KAIST}
    \icmlauthor{SooJean Han}{KAIST}

  \end{icmlauthorlist}

  \icmlaffiliation{KAIST}{School of Electrical Engineering, Korea Advanced Institute of Science and Technology, Daejeon, Korea}

  \icmlcorrespondingauthor{SooJean Han}{soojean@kaist.ac.kr}

  \icmlkeywords{Machine Learning, ICML}

  \vskip 0.3in
]



\printAffiliationsAndNotice{\icmlEqualContribution}

\begin{abstract}
Classifier free guidance is a standard method for conditional sampling in diffusion models, but its sampling rule is not aligned with the objective used in training. This mismatch induces a structural sampling error through the interaction of conditional and unconditional prediction errors.
We analyze this issue by decomposing the sampling error into a base term and a cross term determined by the alignment of the two errors. Based on this analysis we propose CFG with orthogonal error correction (CFG-OEC), a structural modification that reduces the interaction term. For practical settings where ground truth noise is not observable, we introduce a proxy computed from model predictions and a dynamic method that stabilizes correction across diffusion timesteps.
Experiments in a controlled environment validate our theoretical error decomposition and proxy construction. Image generation on Stable Diffusion v1.5 and Stable Diffusion XL show that CFG-OEC improves FID and CLIP scores over CFG and CFG++ across multiple samplers and guidance regimes.
\end{abstract}

\section{Introduction}

Diffusion models~\cite{ho2020, song2021scorebased} generate samples from a data distribution by adding noise to the data and then removing it through a sequence of denoising steps. As research on generative models has advanced, diffusion models have been widely adopted in many domains, including image generation~\cite{saharia2022photorealistic, rombach2022highresolution}, video generation~\cite{ho2022video, chen2024seine}, and robot planning~\cite{carvalho2024motionplanningdiffusion, chi2023diffusionpolicy}. In many applications, it is important to provide additional conditioning information during generation in order to produce samples with desired properties. Such conditioning information can take various forms, such as class labels~\cite{dhariwal2021diffusion}, text descriptions~\cite{zhang2024texttoimage}, or target states~\cite{janner2022planning}. Conditional diffusion models leverage this information to control generation results and to produce semantically consistent samples~\cite{zhang2023adding}.

Representative approaches for incorporating conditioning information in diffusion models include classifier guidance~\cite{dhariwal2021diffusion} and classifier free guidance (CFG)\cite{cfg}. Classifier guidance steers the generation process using signals from a conditional classifier, but it requires training an additional classifier and can suffer from a distribution mismatch between the classifier and the generative model. To alleviate these issues, CFG was proposed to learn conditional and unconditional predictions within a single model and to combine them during sampling, thereby eliminating the need for an external classifier. Due to its simple implementation and strong empirical performance, CFG has become a de facto standard for conditional generation with diffusion models.

Despite its practical success~\cite{cao2024survey}, CFG suffers from a structural mismatch between training and sampling. During training, a single diffusion model is trained with stochastic condition dropout, learning unconditional predictions from condition-dropped inputs with a certain probability and conditional predictions otherwise. At sampling time, however, the CFG guided prediction is formed as a linear combination of the conditional and unconditional outputs. This guided prediction is not the object directly optimized during training, but a sampling-time construction formed from two learned modes of the same model. Consequently, even when the conditional and unconditional predictions are individually accurate, their sampling-time combination can introduce an additional cross-error component that is not directly controlled during training.

This structural mismatch has not been fully addressed in existing analyses of CFG. Prior studies have mainly focused on guidance scale magnitude or distribution-level alignment~\cite{cfg_zero_star_2025,beta_cfg_2025}, rather than on the error structure of the guided predictor itself. To analyze this structural error directly, we study CFG through its deviation from the true conditional score. We explicitly decompose the sampling error into an unconditional prediction error, a conditional prediction error, and an interaction cross term between them. This reveals a structural limitation of CFG. It introduces an additional error, referred to as the cross-error term, that arises only at sampling time. Motivated by this perspective, we propose CFG-OEC (Classifier-Free Guidance with Orthogonal Error Correction), which corrects the resulting structural error and thereby enables more stable and consistent conditional sampling.

In practical image generation tasks, ground-truth noise is unavailable, making direct analysis of CFG error difficult. To address this, we first study CFG in a two-dimensional Gaussian mixture model (2D GMM) setting where the ground truth is accessible. This controlled setting reveals the error structure of CFG and motivates the design of CFG-OEC. We then verify in real image generation tasks that the same insights remain predictive in high-dimensional settings. Experiments show that CFG-OEC provides consistent improvements over standard CFG, particularly in low-guidance regimes, while remaining simple to implement.
\vspace{0.5em}

\section{Related Work}

\subsection{Classifier-Free Guidance in Conditional Diffusion}

CFG is a widely used guidance method for conditional diffusion models. CFG trains a single diffusion model to produce conditional and unconditional predictions and combines the two scores during sampling \cite{dhariwal2021diffusion}. This approach removes the need for an external classifier and provides a simple mechanism for guidance with strong empirical performance.

CFG has been adopted broadly in text-based image generation systems. GLIDE~\cite{nichol2022glidephotorealisticimagegeneration} and DALL·E 2~\cite{ramesh2022hierarchicaltextconditionalimagegeneration} apply conditioning dropout during training to enable CFG and improve alignment between text and images. Latent Diffusion Models combine CFG with cross-attention-based conditioning and lead to practical systems such as Stable Diffusion ~\cite{rombach2022highresolution}. In large-scale models such as Imagen ~\cite{saharia2022photorealistic} CFG is a central role in controlling generation quality and is often used together with additional stabilization techniques during sampling. These works indicate that the linear score combination used by CFG is a core component in modern conditional diffusion systems.
\vspace{1em}

\subsection{Limitations of Existing CFG Corrections}

Despite its empirical success, CFG is not theoretically equivalent to exact conditional sampling. Several approaches have been proposed to mitigate the mismatch. CFG++~\cite{cfgpp} reformulates the guided update under a data manifold constraint to reduce off-manifold updates caused by geometric misalignment between scores. In latent diffusion models, norm amplification effects have been analyzed, and angle domain guidance was proposed to adjust the update direction while controlling its magnitude ~\cite{balaji2022ediffi}. Other works adapt the guidance scale across diffusion timesteps or learn guidance weights with auxiliary models to reduce training sampling mismatch ~\cite{hong2023improving}.

Whereas prior methods mainly adjust the guided update through geometric constraints, norm control, or adaptive scaling, we focus on the error structure induced by the guided combination itself and propose a correction that explicitly reduces the cross term between conditional and unconditional errors.
\vspace{0.5em}

\section{Preliminaries}

\subsection{Diffusion Models}

Diffusion models represent data distributions through a forward noising process and a reverse denoising process.  
In the forward process, Gaussian noise is gradually added to training samples $x_0\sim p_{\text{data}}$ from the original data distribution, producing a noisy sample $x_t\sim p_t$.
Following DDPM, we define the forward noising process at timestep $t$ as
\begin{equation}\label{eq:forward_diffusion}
x_t = \sqrt{\alpha_t}\, x_0 + \sqrt{1 - \alpha_t}\, \epsilon
\end{equation}
where $\epsilon\sim\mathcal{N}(0,I)$ and $\alpha_t\coloneq \prod_{s=1}^t(1 - \beta_s)$ for a fixed variance schedule $\beta_t$.
In the reverse process, a denoising model $\epsilon_\theta$ is trained to remove the injected noise by taking $x_t$ as input and predicting the noise $\epsilon$ used in the forward process.  
The training objective minimizes the mean squared error between the predicted noise and the true noise.

The noise predictor is related to the marginal distribution $p_t(x_t)$ at timestep $t$.
\begin{equation}
\epsilon^*_\theta(x_t)
=
- \sqrt{1 - \alpha_t}\, \nabla_{x_t} \log p_t(x_t).
\end{equation}
For conditional generation, this relation extends to the conditional distribution $p_t(x_t \mid y)$.
\begin{equation}
\epsilon^*_\theta(x_t \mid y)
=- \sqrt{1 - \alpha_t}\, \nabla_{x_t} \log p_t(x_t \mid y).
\end{equation}
These relations show that noise prediction is equivalent to score estimation up to up to a scalar factor determined by the diffusion schedule.

\subsection{Classifier-Free Guidance}
CFG~\cite{cfg} is a standard method for conditional sampling in diffusion models. Its goal is to construct a guided noise predictor
\[
\epsilon_\theta^{(w)}(x_t \mid y)
\]
that steers generation toward the condition \(y\) without requiring an external classifier.

To this end, a single diffusion model is trained to produce both conditional and unconditional noise predictions. During training, the conditioning input \(y\) is randomly dropped, so that the same model learns both \(\epsilon_\theta(x_t \mid y)\) and \(\epsilon_\theta(x_t)\).

At sampling time, CFG combines these two predictions to form a guided predictor. By Bayes' rule,
\[
\nabla_{x_t}\log p_t(y \mid x_t)
=
\nabla_{x_t}\log p_t(x_t \mid y)
-
\nabla_{x_t}\log p_t(x_t),
\]
which shows that the conditional guidance term is given by the difference between the conditional and unconditional scores. Under the standard score-noise correspondence, the guidance term is proportional to
\[
\epsilon_\theta(x_t \mid y) - \epsilon_\theta(x_t).
\]
CFG amplifies this guidance by a scale \(w \ge 0\) and defines the guided prediction as
\begin{equation}\label{eq:cfg_guided_prediction}
\epsilon_\theta^{(w)}(x_t \mid y)
=
w\,\epsilon_\theta(x_t \mid y)
+
(1-w)\,\epsilon_\theta(x_t).
\end{equation}
This guided prediction is then used in the denoising update to perform conditional sampling.

Importantly, \(\epsilon_\theta^{(w)}(x_t \mid y)\) is not itself a training target, but a sampling-time combination of two separately learned predictions. This training-sampling mismatch gives rise to the structural error analyzed in the following section.
\vspace{0.5em}
\begin{figure*}[h]
    \centering
    \begin{subfigure}[b]{0.23\textwidth}
        \centering
        \includegraphics[width=\linewidth]{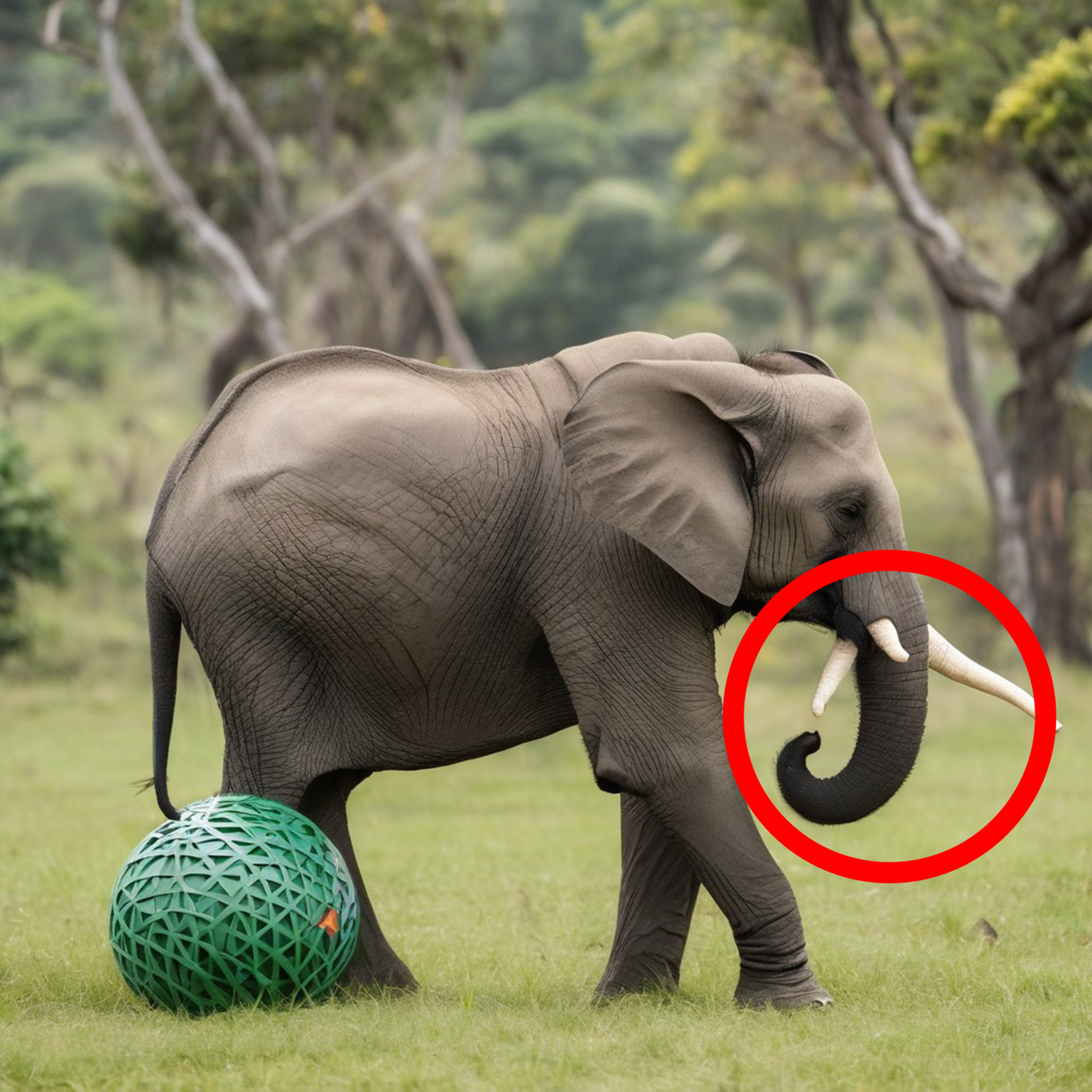}
        \caption*{CFG++}
    \end{subfigure}
    \hfill
    \begin{subfigure}[b]{0.23\textwidth}
        \centering
        \includegraphics[width=\linewidth]{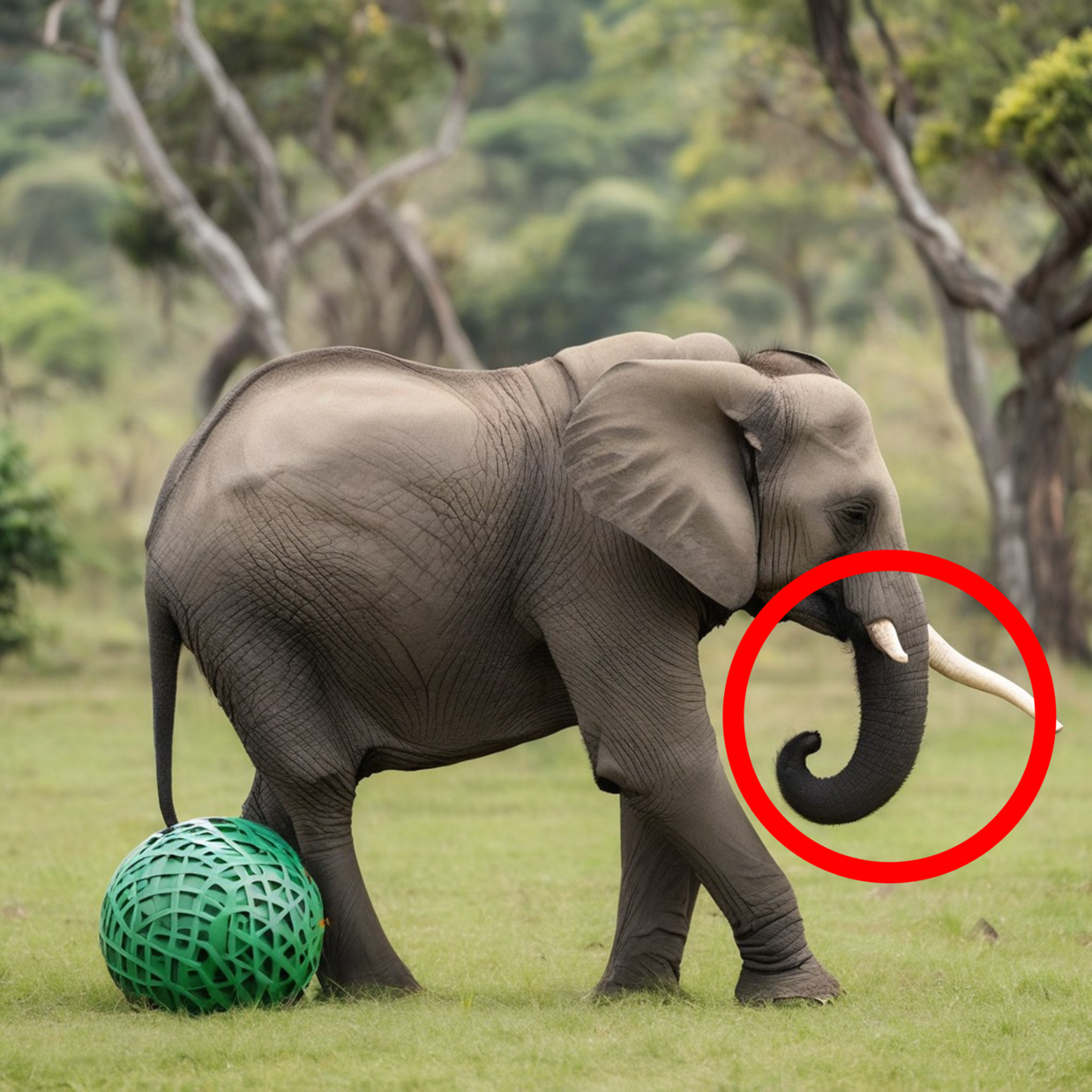}
        \caption*{CFG-OEC}
    \end{subfigure}
    \hfill
    \begin{subfigure}[b]{0.23\textwidth}
        \centering
        \includegraphics[width=\linewidth]{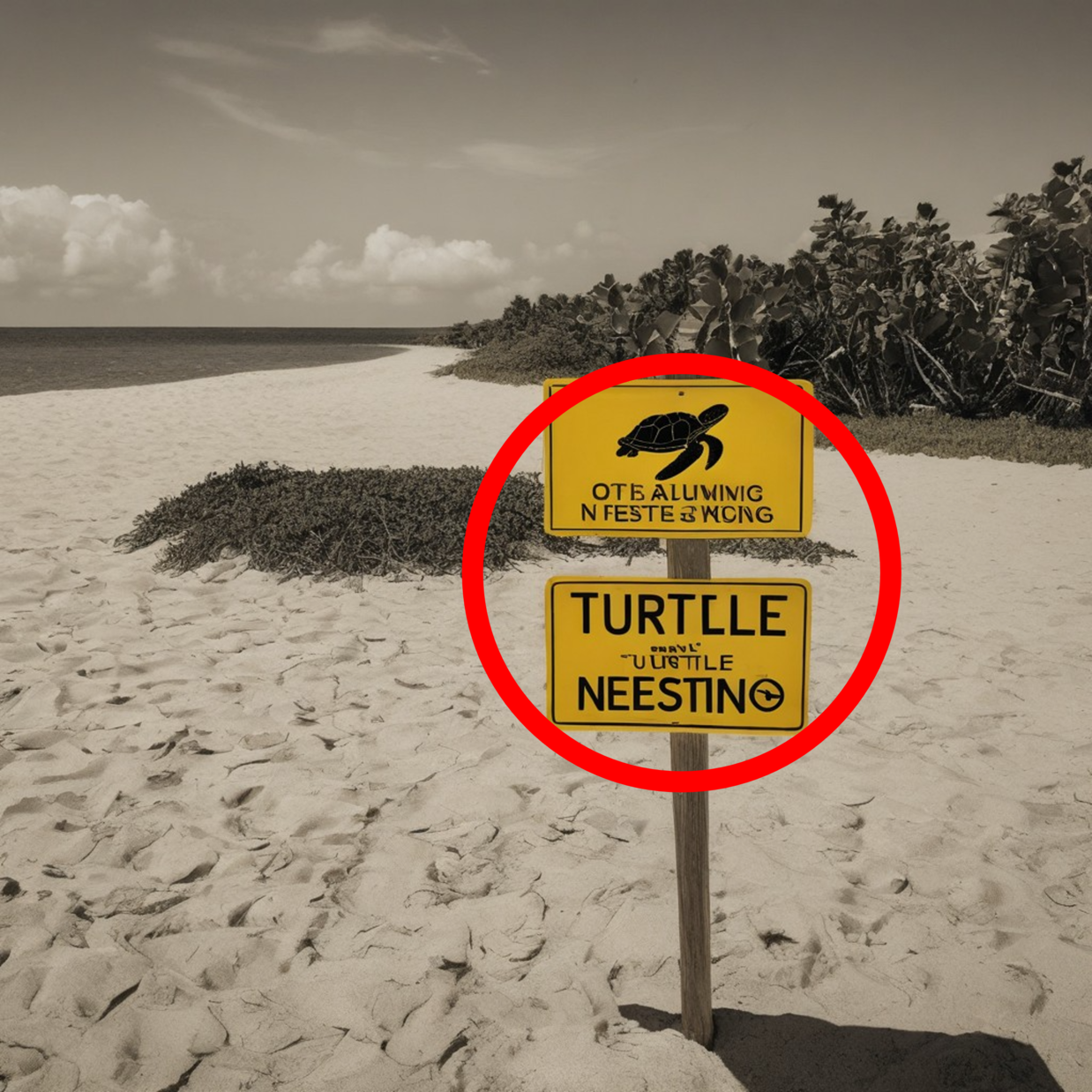}
        \caption*{CFG++}
    \end{subfigure}
    \hfill
    \begin{subfigure}[b]{0.23\textwidth}
        \centering
        \includegraphics[width=\linewidth]{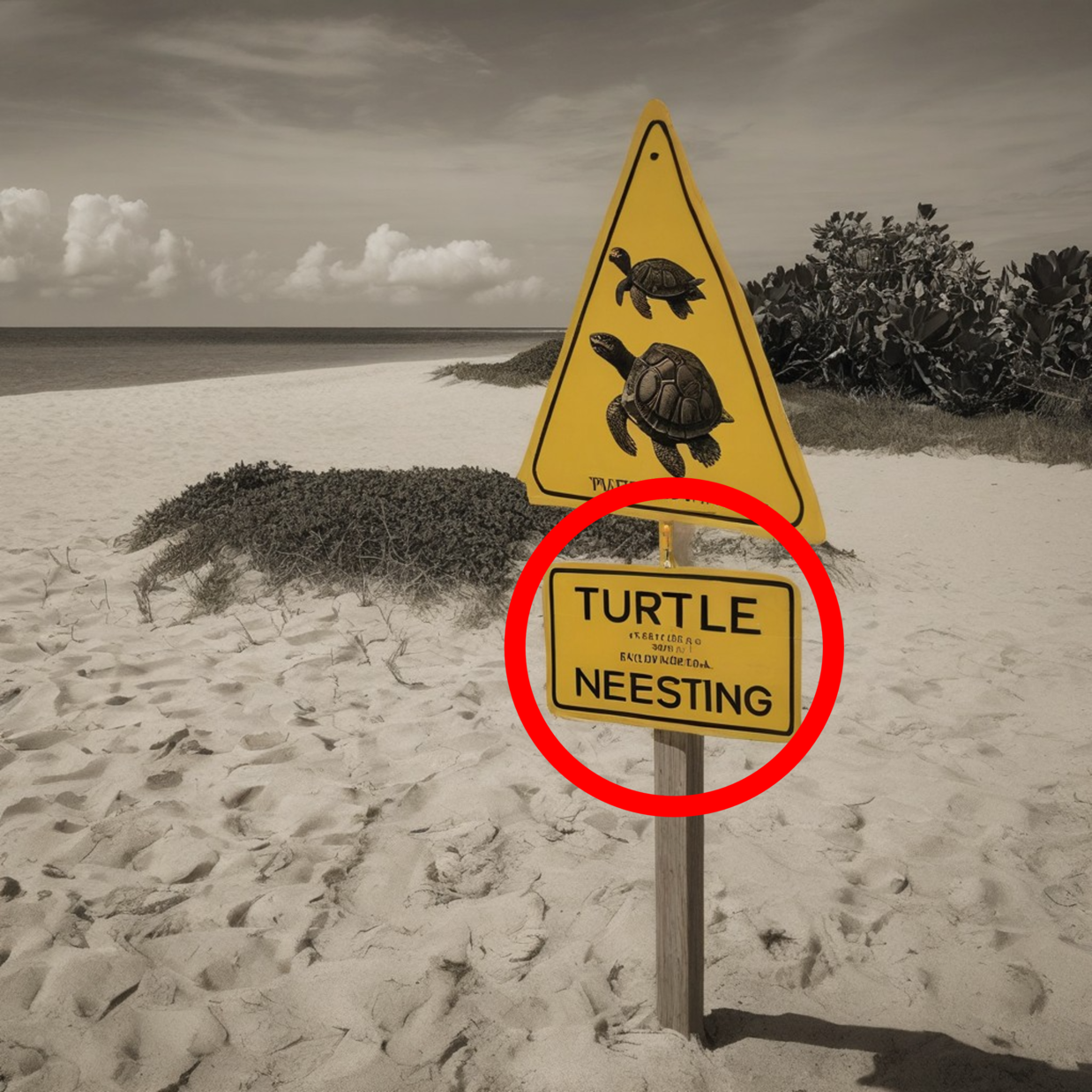}
        \caption*{CFG-OEC}
    \end{subfigure}

    \caption{Comparison of text-to-image(T2I) generation results from Stable Diffusion XL~\cite{stablexl} with DDIM~\cite{ddim} using 50 NFEs. In each example, CFG++ is shown on the left and CFG-OEC on the right. The CFG++ baseline exhibits structural or textual artifacts (e.g., an extra tusk or garbled text), whereas CFG-OEC produces more coherent and visually consistent results.}
    \label{fig:four_across_two_columns}
\end{figure*}

\section{CFG-OEC}

In this section, we introduce \textit{Classifier-Free Guidance with Orthogonal Error Correction} (CFG-OEC), a method designed to mitigate the structural error induced by classifier-free guidance.
We begin by analyzing the error structure of CFG and show that the mismatch between training and sampling gives rise to the cross term. We then present the core idea of CFG-OEC, which reduces this interaction by orthogonalizing the conditional and unconditional error components. Finally, we develop a practical proxy design for real CFG settings, where ground-truth noise is inaccessible. Throughout the section, we use a 2D Gaussian mixture model (2D GMM) as a running example, which enables direct analysis of score functions and distributions~\cite{song2019}.

\subsection{Structural Error in Classifier-Free Guidance}

To formalize the sampling-time cross error in CFG, fix a timestep \(t\), and let \(x_t\) be generated by the forward process in~\eqref{eq:forward_diffusion} from \((x_0,y)\sim p_{\mathrm{data}}\) and \(\epsilon^\ast_t\sim\mathcal N(0,I)\). Define the unconditional and conditional prediction error vectors by
\begin{subequations}
\begin{align}
\Delta_{uc}
&\coloneqq
\epsilon_{\theta}(x_t)
-
\epsilon^\ast_t,\\
\Delta_{c}
&\coloneqq
\epsilon_{\theta}(x_t \mid y)
-
\epsilon^\ast_t.
\label{eq:error_def}
\end{align}
\end{subequations}
At each training step, the conditioning variable $y$ is removed with dropout probability $p \in (0,1)$. The expected training error at timestep $t$ can be written as
\begin{equation}\label{eq:train_error}
\begin{aligned}
e_t^{\text{train}}
\coloneqq&\;
p \,\mathbb{E}_{(x_0,y)\sim p_{\text{data}},\, \epsilon^\ast \sim \mathcal{N}(0,I)}
\left[\left\|
\Delta_{uc}
\right\|^2\right]
\\
&+
(1-p)\,\mathbb{E}_{(x_0,y)\sim p_{\text{data}},\, \epsilon^\ast \sim \mathcal{N}(0,I)}
\left[\left\|
\Delta_{c}
\right\|^2\right].
\end{aligned}
\end{equation}
By contrast, the quantity that matters during CFG sampling is not~\eqref{eq:train_error}. Instead, it is the mean squared error of the guided predictor.
\begin{align}\label{eq:sample_error}
e_t^{\mathrm{sample}}
\coloneqq
\mathbb{E}_{(x_0,y)\sim p_{\text{data}},\, \epsilon^\ast \sim \mathcal{N}(0,I)}
\big[\| \omega \Delta_{c} + (1-\omega)\Delta_{uc}\|^2\big].
\end{align}
This expression can be expanded as
\begin{equation}\label{eq:sample_error_2}
\begin{aligned}
e^{\mathrm{sample}}_t &= \underbrace{
\mathbb{E}\left[\omega^2\|\Delta_{c}\|^2 + (1-\omega)^2\|\Delta_{uc}\|^2
\right]}_{\text{Base Error}}\\
&+
\underbrace{\mathbb{E}\left[
2\omega(1-\omega)
\langle
\Delta_{c},
\Delta_{uc}\rangle\right]
}_{\text{Cross Error}} .
\end{aligned}
\end{equation}

The base error is determined by the conditional and unconditional prediction errors, both of which are penalized during training. However, the cross error depends on the relative alignment between the conditional and unconditional error vectors and is not explicitly penalized by the training objective. As a result, minimizing the objective does not in itself control the sampling-time error induced by their interaction.

\vspace{0.5em}
\begin{table}[h]
\centering
\setlength{\tabcolsep}{4pt}
\renewcommand{\arraystretch}{1.0}
\begin{tabular}{cccc}
\hline
$\omega$ & Base Error & Cross Error & Ratio \\
\hline
0.5 & $0.0058 \pm 0.0004$ & $0.0006 \pm 0.0001$ & $0.1057$ \\
0.8 & $0.0033 \pm 0.0001$ & $0.0004 \pm 0.0000$ & $0.1200$ \\
3.0 & $0.1126 \pm 0.0051$ & $-0.0147 \pm 0.0014$ & $-0.1309$ \\
5.0 & $0.4075 \pm 0.0236$ & $-0.0489 \pm 0.0045$ & $-0.1206$ \\
7.0 & $0.8887 \pm 0.0554$ & $-0.1027 \pm 0.0095$ & $-0.1162$ \\
9.0 & $1.5562 \pm 0.1003$ & $-0.1761 \pm 0.0162$ & $-0.1138$ \\
\hline
\end{tabular}
\caption{Base error, Cross error and Ratio (cross error / base error) of CFG measured in a 2D GMM environment.}
\label{tab:cfg_ec_ratio_mix}
\end{table}

\begin{example}\label{ex:2d_gmm_ex1}
To illustrate this structural mismatch, we analyze the base and cross errors in a 2D GMM. Table~\ref{tab:cfg_ec_ratio_mix} reports their magnitudes and relative ratios as functions of \(\omega\). The detailed experimental setup is provided in Appendix~\ref{sec:toy_experiment_setup}.
Consistent with the error decomposition, we observe that for $\omega \in (0,1)$ the cross error increases the total sampling error. Interestingly, for \(\omega > 1\), the cross term in the decomposition often becomes negative and partially offsets the base error. This indicates that the sampling-time interaction can reduce the total error incidentally, although this effect is neither controlled nor guaranteed by training. 
In the \(\omega \in (0,1)\) range, the cross error accounts for approximately \(10\%\) to \(15\%\) of the base error. This shows that the cross error is not merely a minor residual effect, but a non-trivial structural component of the guided denoising error in CFG.
\null\hfill\qedsymbol
\end{example}

\subsection{Orthogonal Error Correction}

We introduce orthogonal error correction, which reduces the structural error of CFG by making the unconditional error vector orthogonal to the conditional error vector. This eliminates the cross term while leaving the conditional prediction unchanged, thereby preserving the conditional information required for the task.

The orthogonalized unconditional error vector \(\Delta_{uc}^{\perp}\) is obtained by removing from \(\Delta_{uc}\) its component parallel to \(\Delta_c\):
\begin{equation}
\Delta_{uc}^{\perp}
=
\Delta_{uc}
-
\mathrm{proj}_{\Delta_c}(\Delta_{uc}),
\label{eq:true_orthogonalization}
\end{equation}
where \(\mathrm{proj}_{\Delta_c}(\Delta_{uc})\) denotes the projection of \(\Delta_{uc}\) onto \(\Delta_c\).

Substituting \(\Delta_{uc}^{\perp}\) in place of \(\Delta_{uc}\) in Eq.~\eqref{eq:sample_error_2} removes the cross term and yields
\[
e_t^{\mathrm{sample},\perp}
=
\mathbb{E}\!\left[
\omega^2\|\Delta_c\|^2 + (1-\omega)^2\|\Delta_{uc}^{\perp}\|^2
\right].
\]
Since orthogonal projection does not increase norm, \(\|\Delta_{uc}^{\perp}\|^2 \le \|\Delta_{uc}\|^2\). Therefore,
\[
e_t^{\mathrm{sample},\perp}
\le
\mathbb{E}\!\left[
\omega^2\|\Delta_c\|^2 + (1-\omega)^2\|\Delta_{uc}\|^2
\right]
=
e_t^{\mathrm{base}},
\]
where \(e_t^{\mathrm{base}}\) denotes the base term in the decomposition of Eq.~\eqref{eq:sample_error_2}. Thus, orthogonal error correction removes the cross term while not increasing the base component of the sampling error.

\begin{table}[h]
\centering
\setlength{\tabcolsep}{8pt}
\renewcommand{\arraystretch}{1.1}
\begin{tabular}{ccc}
\hline
$\omega$ & $B\%$ & $T\%$ \\
\hline
0.5 & $-41.7053 \pm 0.9171$ & $-47.2778 \pm 0.7009$ \\
0.8 & $-11.9605 \pm 1.5076$ & $-21.3910 \pm 1.3799$ \\
3.0 & $-34.4968 \pm 1.6374$ & $-24.6000 \pm 3.0491$ \\
5.0 & $-38.1162 \pm 1.3182$ & $-29.5963 \pm 2.5885$ \\
7.0 & $-39.3269 \pm 1.1914$ & $-31.3203 \pm 2.3935$ \\
9.0 & $-39.9287 \pm 1.1250$ & $-32.1876 \pm 2.2888$ \\
\hline
\end{tabular}
\caption{
Relative changes of base error ($B\%$) and total error ($T\%$)
with respect to the guidance scale $\omega$ in a controlled 2D GMM
CFG experiment (mean $\pm$ std over seeds).
}
\label{tab:summary_ratio_only_w}
\end{table}

\begin{example}\label{ex:2d_gmm_ex2}
We evaluate orthogonal error correction in the setting of Example~\ref{ex:2d_gmm_ex1}. Table~\ref{tab:summary_ratio_only_w} reports the relative changes in the base error (B\%) and total sampling error (T\%) after correction under different guidance scales, where negative values indicate error reduction.

For \(0<\omega<1\), orthogonal correction reduces both the base error and the total sampling error. In this regime, the removal of the cross term appears to contribute to the reduction in total sampling error after correction.

For \(\omega>1\), the cross term may decrease the total sampling error before correction. Nevertheless, the reduction in the base error remains larger in magnitude, so the total sampling error still decreases after correction. In this 2D GMM example, orthogonal error correction consistently reduces the total sampling error across the tested guidance regimes.\null\hfill\qedsymbol
\end{example}

\subsection{Proxy Design for Error Correction}
Orthogonal error correction in the previous subsection assumes access to the ground-truth noise \(\epsilon_t^\ast\), which is unavailable in practice. We therefore replace \(\epsilon_t^\ast\) with a proxy constructed from adjacent model predictions along the sampling trajectory. Let \(\epsilon_\theta(x_t)\) denote the noise prediction at timestep \(t\). We define the proxy by linear extrapolation:
\begin{equation}\label{eq:unconditional_proxy}
\hat{\epsilon}_t^{(uc)}
\coloneqq
2\,\epsilon_\theta(x_t)-\epsilon_\theta(x_{t+1}).
\end{equation}
Likewise, using the conditional prediction \(\epsilon_\theta(x_t\mid y)\), we define the conditional proxy as
\begin{equation}\label{eq:conditional_proxy}
\hat{\epsilon}_t^{(c)}
\coloneqq
2\,\epsilon_\theta(x_t\mid y)-\epsilon_\theta(x_{t+1}\mid y).
\end{equation}
In both cases, the proxy uses the local change between successive predictions to extrapolate a nearby directional trend along the sampling trajectory. We do not interpret \(\hat{\epsilon}_t^{(uc)}\) or \(\hat{\epsilon}_t^{(c)}\) as exact estimates of the true noise target. Rather, they serve as computable surrogates that may preserve useful local directional information for the correction step.

Using the extrapolated proxies, we define the proxy-based error vectors
\begin{equation}
A = \epsilon_\theta(x_t \mid y) - \hat{\epsilon}_t^{(c)}, \qquad
B = \epsilon_\theta(x_t) - \hat{\epsilon}_t^{(uc)}.
\end{equation}
which approximate the conditional and unconditional error vectors, respectively. We then apply a proxy-based orthogonal error correction to remove from the unconditional error the component aligned with the conditional error:
\begin{equation}
\label{eq:proxy_orthogonalization}
B_{\perp}
=
B
-
\frac{\langle A, B \rangle}{\|A\|^2}A.
\end{equation}
The corrected unconditional prediction is reconstructed as
\begin{equation}
\bar{\epsilon}_t^{(uc)}
=
B_{\perp}
+
\hat{\epsilon}_t^{(uc)},
\end{equation}
thereby removing the aligned component that contributes to the cross-error interaction while retaining the remaining unconditional error structure.

To improve robustness, we further introduce a dynamic mixing step. Rather than using \(\bar{\epsilon}_t^{(uc)}\) directly, we define the final unconditional prediction as
\begin{equation}
\tilde{\epsilon}_t^{(uc)}
=
(1-s)\,\bar{\epsilon}_t^{(uc)}
+
s\,\epsilon_\theta(x_t)^{(uc)},
\end{equation}
where \(s \in [0,1]\) is computed from the cosine similarity between \(A\) and \(B\). Since the cross-error term in \eqref{eq:sample_error_2} depends on the alignment between the conditional and unconditional error directions, larger values of \(s\) indicate stronger alignment between the two proxy-induced differences. In this regime, the projection step removes a larger component of the unconditional proxy-induced difference, making the correction more sensitive to proxy inaccuracy and therefore potentially less stable. By contrast, when \(s\) is smaller, the alignment is weaker, and the corrected prediction can be incorporated more reliably. Accordingly, we apply the correction only when \(s < \tau\). The complete CFG-OEC procedure is summarized in algorithm~\ref{alg:our_algorithm_math}.

\vspace{0.5em}

\begin{example}
We evaluate CFG-OEC in the setting of Example~\ref{ex:2d_gmm_ex1} without access to the ground-truth noise \(\epsilon^\ast\), using the proposed proxy with dynamic mixing. In this controlled setting, we measure the cosine similarity between the proxy and the ground-truth noise to assess whether the proxy preserves useful directional information for the correction step.

\begin{figure}[h]
\centering
\includegraphics[width=1.0\linewidth]{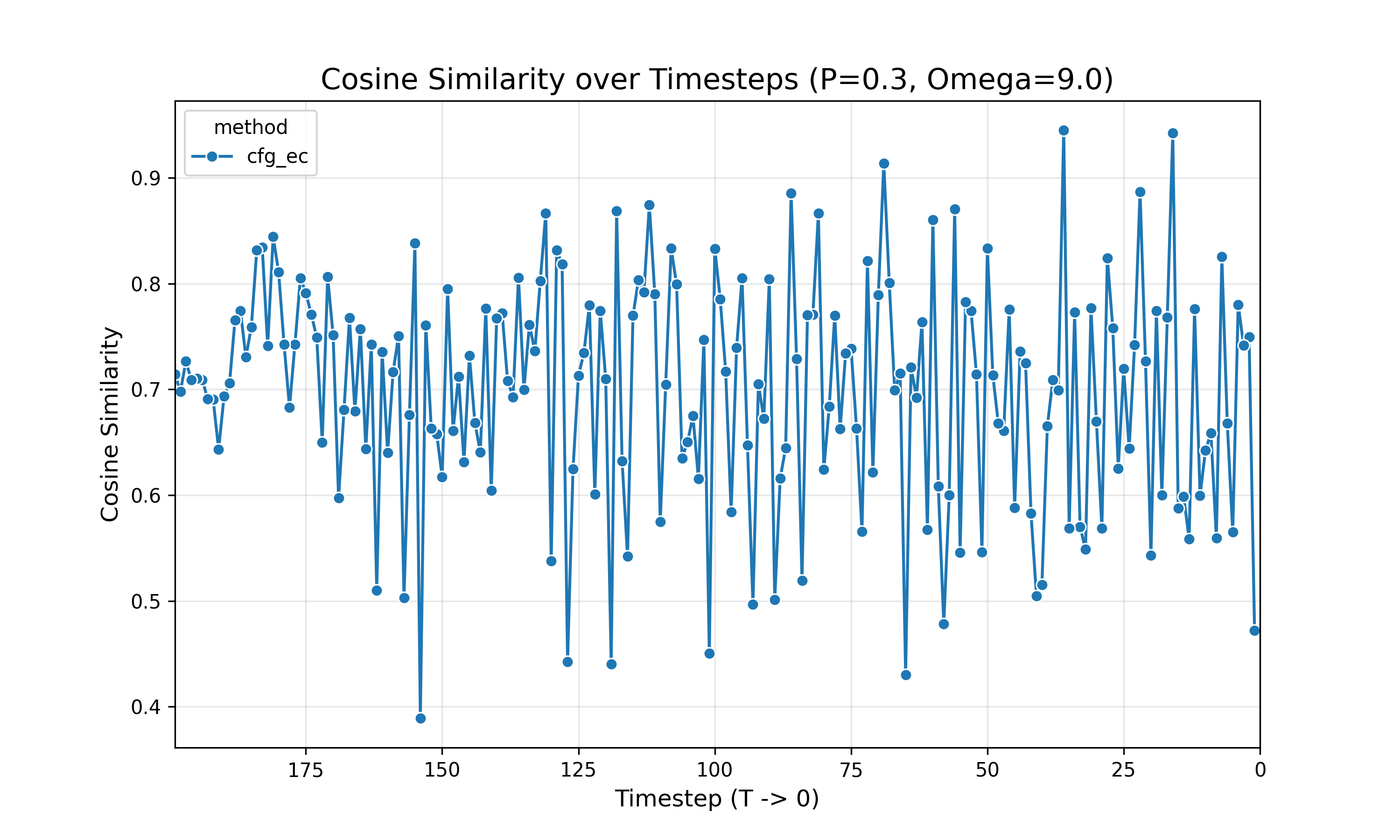}
\caption{
Cosine similarity between the proxy and the ground-truth noise across guidance scales \(\omega\) in the 2D GMM environment. 
}
\label{fig:proxy_cosine_similarity}
\end{figure}
Figure~\ref{fig:proxy_cosine_similarity} shows that the proposed proxy remains well aligned with the ground-truth noise across diffusion timesteps, despite being constructed only from adjacent-timestep predictions. This suggests that the proxy captures local directional trends without direct access to the true noise target, although it does not directly validate recovery of the \(\epsilon^\ast_t\).
\end{example}

\begin{figure} 
    \centering
    \begin{subfigure}[b]{0.48\linewidth}
        \centering
        \includegraphics[width=\linewidth]{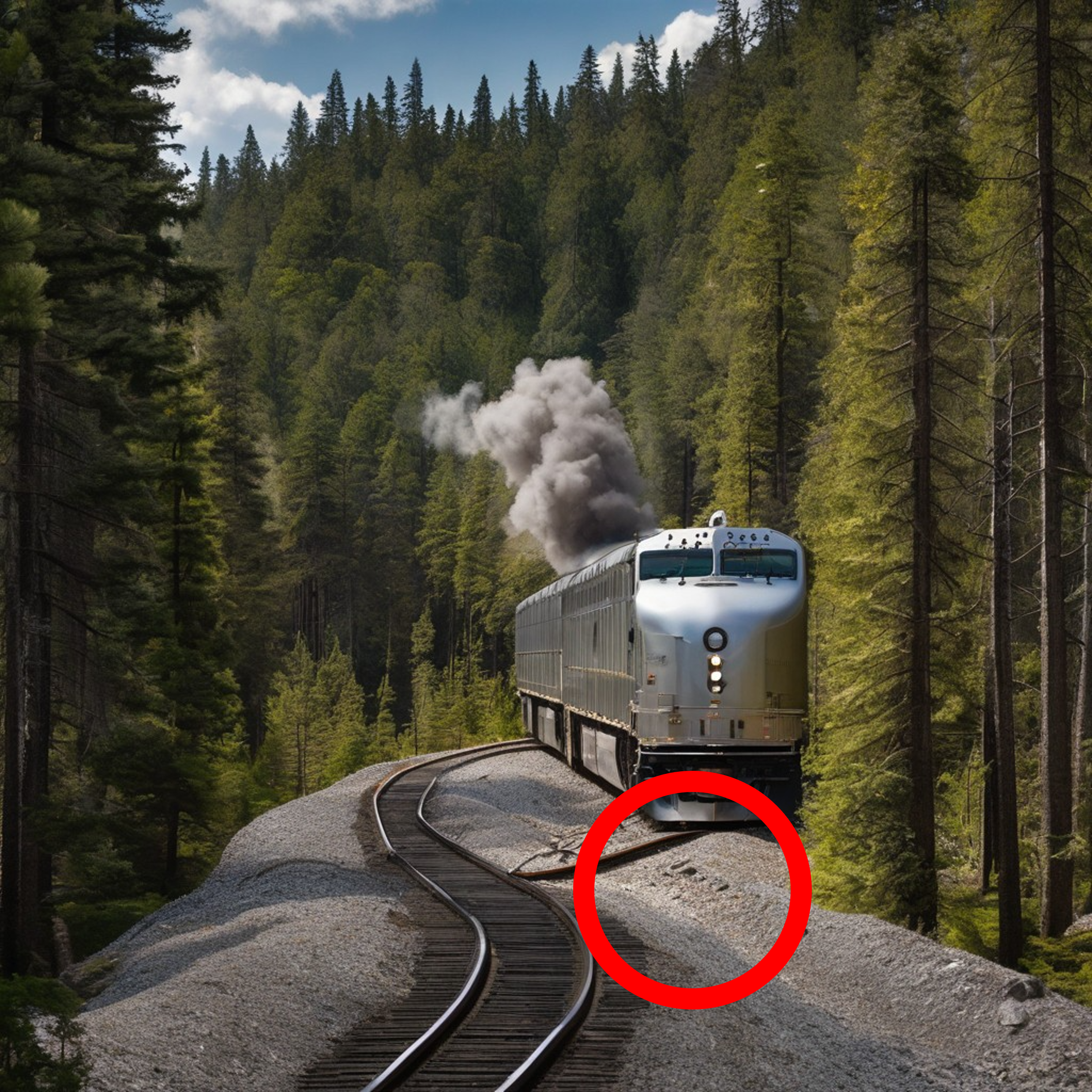}
        \caption{CFG++}
        \label{fig:cfg_plus_plus_1}
    \end{subfigure}
    \hfill
    \begin{subfigure}[b]{0.48\linewidth}
        \centering
        \includegraphics[width=\linewidth]{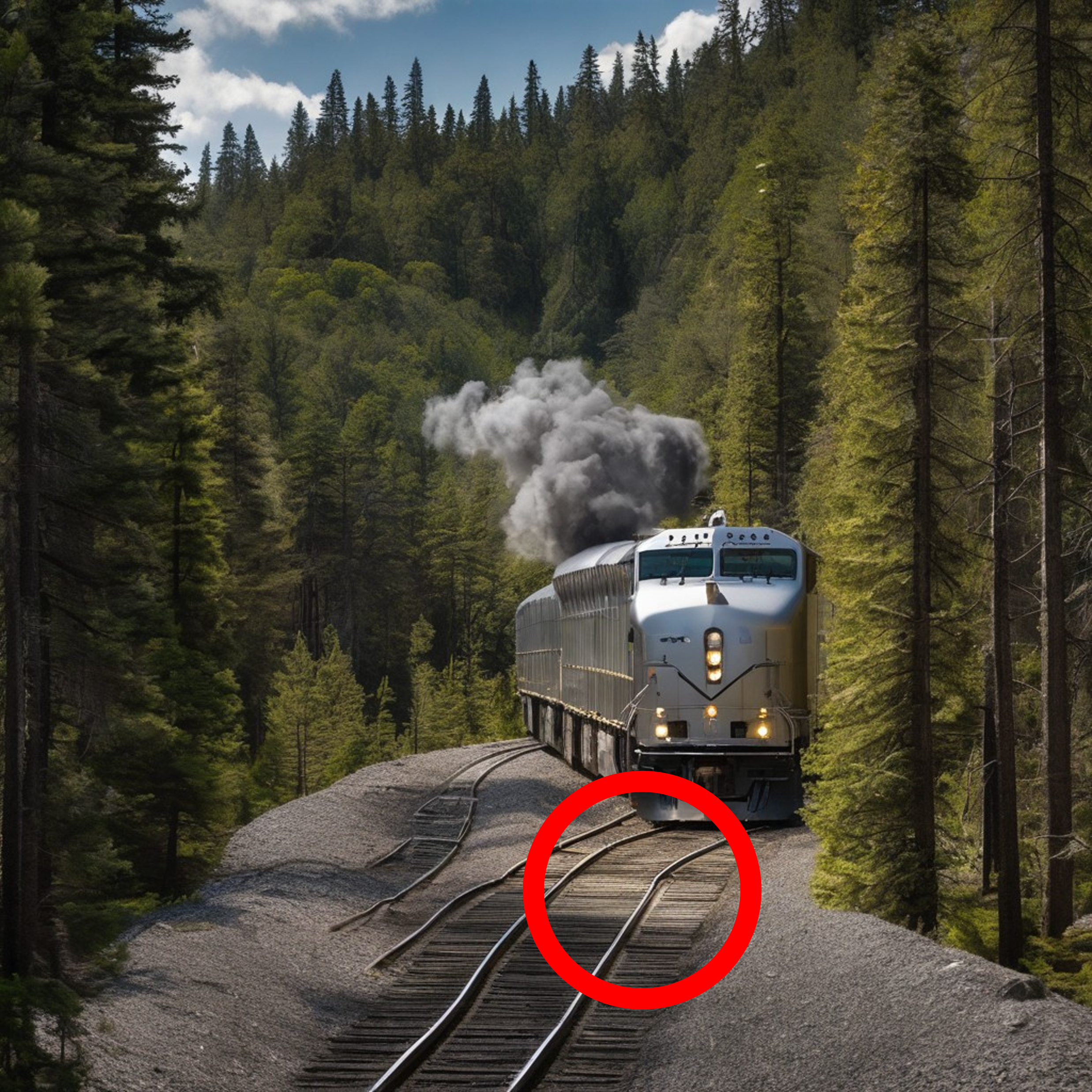}
        \caption{Ours}
        \label{fig:ours_1}
    \end{subfigure}

    \caption{Comparison of SDXL T2I generation with DDIM using 50 NFEs.
    The baseline (CFG++) exhibits a structural artifact with a physically incorrect railway track,
    whereas our method produces a coherent structure.}
\end{figure}

\begin{table*}[t] 
  \centering 
  \setlength{\tabcolsep}{3.0pt} 
  \begin{tabular}{l cc cc cc cc cc}
    \toprule
    & \multicolumn{2}{c}{$\omega$ = 0.5} &  \multicolumn{2}{c}{$\omega$ = 2.5} & \multicolumn{2}{c}{ $\omega$ = 5.0} & \multicolumn{2}{c}{ $\omega$ = 7.5}\\
    \cmidrule(lr){2-3} \cmidrule(lr){4-5} \cmidrule(lr){6-7} \cmidrule(lr){8-9}  
    Method & FID $\downarrow$ & CLIP $\uparrow$ & FID $\downarrow$ & CLIP $\uparrow$ & FID $\downarrow$ & CLIP $\uparrow$ & FID $\downarrow$ & CLIP $\uparrow$ \\
    \midrule
    CFG~\cite{cfg} & 64.74 & 21.23  & \textbf{13.18} & 31.05 & 15.81 & 31.75 & 18.62 & 31.87\\
    CFG-OEC & \textbf{62.52} & \textbf{21.50}  & 13.23 & \textbf{31.09} & \textbf{15.77} & \textbf{31.86} & \textbf{18.59} & \textbf{31.96} \\
    \bottomrule
  \end{tabular}
  \caption{Quantitative evaluation of the DPM-Solver++(2M) 20 NFEs on MSCOCO 10k dataset.}
  \label{tab:quantitative_results_1} 
\end{table*}

\begin{table*}
  \centering 
  \setlength{\tabcolsep}{3.0pt} 
  \begin{tabular}{l cc cc cc cc cc}
    \toprule
    & \multicolumn{2}{c}{$\omega$ = 0.2} & \multicolumn{2}{c}{$\omega$ = 0.4} & \multicolumn{2}{c}{ $\omega$ = 0.6} & \multicolumn{2}{c}{$\omega$ = 0.8} & \multicolumn{2}{c}{$\omega$ = 1.0} \\
    \cmidrule(lr){2-3} \cmidrule(lr){4-5} \cmidrule(lr){6-7} \cmidrule(lr){8-9}  \cmidrule(lr){10-11}
    Method & FID $\downarrow$ & CLIP $\uparrow$ & FID $\downarrow$ & CLIP $\uparrow$ & FID $\downarrow$ & CLIP $\uparrow$ & FID $\downarrow$ & CLIP $\uparrow$ & FID $\downarrow$ & CLIP $\uparrow$ \\
    \midrule
    CFG++~\cite{cfgpp} & 13.19 & 31.15 & 15.94 & 31.79 & 18.81 & 31.95 & \textbf{20.74} & 31.99 & \textbf{22.18} & 31.99\\
    CFG-OEC++  & \textbf{13.18} & \textbf{31.18} & \textbf{15.82} & \textbf{31.85} & \textbf{18.73} & \textbf{32.01} & 20.76 & \textbf{32.06} & 22.25 & \textbf{32.05} \\
    \bottomrule
  \end{tabular}
  \caption{Quantitative evaluation of 50 NFEs DDIM T2I with SD v1.5 on MSCOCO 10k dataset.}
  \label{tab:quantitative_results_2} 
\end{table*}

\section{Experiment}

\subsection{Experimental Settings}

We conduct experiments on Stable Diffusion XL~\cite{stablexl} and Stable Diffusion v1.5~\cite{rombach2022highresolution}. We use DDIM~\cite{ddim}, UniPC~\cite{unipc}, and DPM-Solver++(2M)~\cite{dpm_solver_plus} as samplers. DDIM is run with 50 NFEs, while UniPC and DPM-Solver++(2M) use 20 NFEs. We compare against CFG and CFG++ as baselines. Text prompts are taken from the MSCOCO caption~\cite{lin2014coco_simple} dataset, using the first caption for each image. We evaluate image quality and text alignment using FID~\cite{fid} and CLIP~\cite{clip}. Unless otherwise stated, all reported results use the dynamic-mixing version of CFG-OEC.

\subsection{Results}

We first evaluate CFG-OEC across guidance scales on the MSCOCO 10k validation set using DPM-Solver++(2M) with 20 NFEs and the Karras sigma schedule~\cite{Karras2022EDM}.
The results in \cref{tab:quantitative_results_1} show that CFG-OEC improves image generation performance across a range of guidance scales, with the clearest gains in the low-guidance regime. At \(\omega=0.5\), CFG-OEC yields better quantitative performance than CFG in terms of both FID and CLIP. At higher guidance scales, the method remains competitive. In particular, at \(\omega=5.0\) and \(\omega=7.5\), the proposed method improves both FID and CLIP, suggesting gains in both visual fidelity and text alignment. At \(\omega=2.5\), the method achieves a slightly higher CLIP score with a small increase in FID, indicating a mild fidelity-alignment trade-off in that regime. Overall, the results show that the proposed correction remains effective across different guidance regimes, while being beneficial at lower guidance scales.

CFG++ is designed to improve low-guidance sampling by making better use of unconditional estimates than standard CFG. To examine whether the proposed correction extends beyond the original CFG setting, we combine it with CFG++ and denote the resulting method by CFG-OEC++. We evaluate this variant on Stable Diffusion v1.5 with a 50-step DDIM sampler on the MSCOCO 10k validation set.

As shown in \cref{tab:quantitative_results_2}, CFG-OEC++ improves both FID and CLIP over CFG++ in three of the five tested configurations (\(\omega=0.2,0.4,0.6\)). The clearest gain appears at \(\omega=0.4\). At higher guidance scales (\(\omega=0.8,1.0\)), the method yields slightly higher CLIP scores with a small degradation in FID, indicating a mild fidelity-alignment trade-off in that regime.
The gains are distinct in \(\omega=0.4\) and \(0.6\), which is consistent with our theory, since the coefficient \(2\omega(1-\omega)\) is largest around \(\omega=0.5\) on \([0,1]\). The empirical trend, supports the relevance of the proposed method.

\begin{table}
\centering
\resizebox{1.0\columnwidth}{!}{
    \begin{tabular}{llcc}
    \toprule
    \ Model & Method (Guidance Scale) & FID (↓) & CLIP (↑) \\
    \midrule
    \midrule
    SD v1.5 (DPM++ 2M) & CFG++ (0.5) & 13.15 & 30.85 \\
    SD v1.5 (DPM++ 2M) & CFG-OEC++ (0.5) & \textbf{13.14} & \textbf{30.91} \\
    \midrule
    SDXL (DDIM) & CFG++ (0.5) & 23.17 & 31.87 \\
    SDXL (DDIM) & CFG-OEC++ (0.5) & \textbf{23.11} & \textbf{31.95} \\
    \bottomrule                     
    \end{tabular}
}
\caption{Quantitative comparison on the MSCOCO dataset on T2I sampling}
\label{tab:quantitative_results_3}
\end{table}

As presented in \cref{tab:quantitative_results_3}, the proposed correction transfers across additional sampler-model combinations. In both the SD v1.5 + DPM-Solver++(2M) setting and the SDXL + DDIM setting, CFG-OEC++ improves FID and CLIP over CFG++. These results suggest that the proposed correction is compatible with multiple diffusion models and samplers.

\subsection{Ablation Study}

To validate the efficacy and necessity of the dynamic mixing strategy, we conducted an ablation study across three distinct variants of our method. Each variant differs in how it combines the original unconditional prediction with the orthogonally corrected one.

The compared variants are as follows.\\
\textbf{CFG-OEC++ (Full)} uses only the fully corrected unconditional prediction. \\
\textbf{CFG-OEC++ (0.7)} uses a fixed mixing ratio between the corrected and original predictions, with weights of \(0.3\) and \(0.7\), respectively. \\
\textbf{CFG-OEC++ (Dynamic)} adaptively determines the mixing ratio based on the cosine similarity between the two proxy-induced differences.

\begin{table}
\centering
\label{tab: necessity of dynamic mixing ratio_results}
\resizebox{0.8\columnwidth}{!}{%
    \begin{tabular}{lcc}
    \toprule
    \textbf{Method} & \textbf{FID Score (↓)} & \textbf{CLIP Score (↑)} \\
    \midrule
    CFG-OEC++ (Full) & 19.04 & \textbf{32.25} \\
    CFG-OEC++ (0.7)      & 18.78 & 32.04 \\
    \textbf{CFG-OEC++ (Dynamic)}      & \textbf{18.73} & 32.01 \\
    \bottomrule
    \end{tabular}%
}
\caption{Quantitative comparison on the MSCOCO 10k on SD v1.5 in 0.6 guidance scale with the DDIM 50 NFEs}
\label{tab:quantitative_results_4}
\end{table}

The results in  \cref{tab:quantitative_results_4} reveal a trade-off between prompt alignment and image fidelity. The Full orthogonalization strategy attains the highest CLIP (32.25). However, this full method comes at the cost of image fidelity, yielding the lowest FID (19.04). Conversely, the Fixed-Ratio Mixing strategy improves the FID (18.78), but this static compromise dilutes the guidance signal, resulting in a lower CLIP. The proposed Dynamic method achieves the best FID (18.73), while simultaneously maintaining a competitive CLIP (32.01) that is nearly on par with the fixed-ratio approach. \\
As shown in \cref{tab:quantitative_results_4}, the proposed dynamic strategy achieves the best FID while maintaining a competitive CLIP score, suggesting that adaptive mixing provides the most favorable fidelity-alignment balance among the tested variants.

\section{Conclusion}

This paper analyzes the structural mismatch between training and sampling in CFG, characterizes how the discrepancy induced by this mismatch behaves across guidance scales, and proposes CFG-OEC as an error correction method to mitigate the resulting structural error. We decompose the sampling error of CFG into a base error and a cross error, and show that the contribution of the cross error depends on the guidance scale while not being explicitly controlled by the standard CFG training objective. We address this issue with an orthogonal error correction strategy that removes the interaction term. For practical settings where ground-truth noise is inaccessible, we further develop a prediction-based proxy and a dynamic stabilization method for timestep-wise error correction. Experimental results show that CFG-OEC behaves consistently with the theoretical analysis in a controlled 2D GMM environment. In large-scale image generation experiments using Stable Diffusion v1.5 and SDXL, CFG-OEC achieves consistent performance improvements over both CFG and CFG++. In particular, the method improves image fidelity and text alignment in low-guidance regimes. 
In higher-guidance settings, the method remains competitive and sometimes exhibits a mild fidelity--alignment trade-off. These results suggest that CFG-OEC provides a practical and theoretically motivated way to mitigate a structural source of error in classifier-free guidance.

\vspace{3em}

\bibliography{example_paper}
\bibliographystyle{icml2026}

\newpage
\appendix
\onecolumn
\section{Pseudo Code}
This section presents the pseudocode of the proposed method.
Algorithm~\cref{alg:our_algorithm_math} describes the sampling procedure of the dynamic CFG-OEC method.

The full method applies the corrected unconditional noise prediction directly, without interpolating it with the original unconditional prediction from the model.

\begin{algorithm}[h]
\caption{CFG-OEC with Proxy-based Dynamic Error Correction}
\label{alg:our_algorithm_math}
\textbf{Input}: Initial noise $x_T$, guidance scale $w$, alignment threshold $\tau$

\begin{algorithmic}[1]
\FOR{$t = T \to 1$}
    
    \STATE Compute conditional and unconditional predictions:
    \STATE $\epsilon_t^{(c)} \leftarrow \epsilon_\theta(x_t \mid y)$
    \STATE $\epsilon_t^{(uc)} \leftarrow \epsilon_\theta(x_t)$

    \IF{$t < T$}
        \STATE // Step 1: Construct proxies via extrapolation
        \STATE $\hat{\epsilon}_t^{(c)} \leftarrow 2\epsilon_t^{(c)} - \epsilon_{t+1}^{(c)}$
        \STATE $\hat{\epsilon}_t^{(uc)} \leftarrow 2\epsilon_t^{(uc)} - \epsilon_{t+1}^{(uc)}$

        \STATE // Step 2: Define proxy-based error vectors
        \STATE $A \leftarrow \epsilon_t^{(c)} - \hat{\epsilon}_t^{(c)}$
        \STATE $B \leftarrow \epsilon_t^{(uc)} - \hat{\epsilon}_t^{(uc)}$

        \STATE // Step 3: Compute alignment (cosine similarity)
        \STATE $s \leftarrow \frac{\langle A, B \rangle}{\|A\| \|B\|}$

        \IF{$s < \tau$}
            \STATE // Step 4: Orthogonalize unconditional error
            \STATE $B_{\perp} \leftarrow B - \frac{\langle A, B \rangle}{\|A\|^2} A$

            \STATE // Step 5: Reconstruct corrected unconditional prediction
            \STATE $\tilde{\epsilon}_t^{(uc)} \leftarrow B_{\perp} + \hat{\epsilon}_t^{(uc)}$

            \STATE // Step 6: Temporal smoothing (dynamic mixing)
            \STATE $\tilde{\epsilon}_t^{(uc)} \leftarrow (1-s)\tilde{\epsilon}_t^{(uc)} + s{\epsilon}_{t}^{(uc)}$
        \ELSE
            \STATE $\tilde{\epsilon}_t^{(uc)} \leftarrow \epsilon_t^{(uc)}$
        \ENDIF
    \ELSE
        \STATE $\tilde{\epsilon}_t^{(uc)} \leftarrow \epsilon_t^{(uc)}$
    \ENDIF

    \STATE // Step 7: CFG prediction
    \STATE $\epsilon_t^{(pred)} \leftarrow \tilde{\epsilon}_t^{(uc)} + w\left(\epsilon_t^{(c)} - \tilde{\epsilon}_t^{(uc)}\right)$

    \STATE // Step 8: Update state
    \STATE $x_{t-1} \leftarrow \mathrm{Solver}(x_t, \epsilon_t^{(pred)})$

\ENDFOR

\STATE \textbf{return} $x_0$
\end{algorithmic}
\end{algorithm}

To maintain temporal consistency and enable stable proxy construction, we reuse the original unconditional prediction at each timestep as the reference value for the subsequent step. However, Cosine similarity lies in [-1,1] in general. In practice, we observe that the values are predominantly nonnegative, and we therefore use it directly as a mixing weight without additional normalization.
\clearpage
\section{Toy Experiment Setup}
\label{sec:toy_experiment_setup}
This paper designs a controlled toy diffusion environment to analyze the structural properties of CFG and CFG-OEC. Instead of using real image data, the environment is constructed to explicitly separate conditional and unconditional prediction errors and to enable quantitative analysis of their interaction.
\begin{figure}[t]
\centering
\begin{subfigure}{0.45\textwidth}
\includegraphics[width=\linewidth]{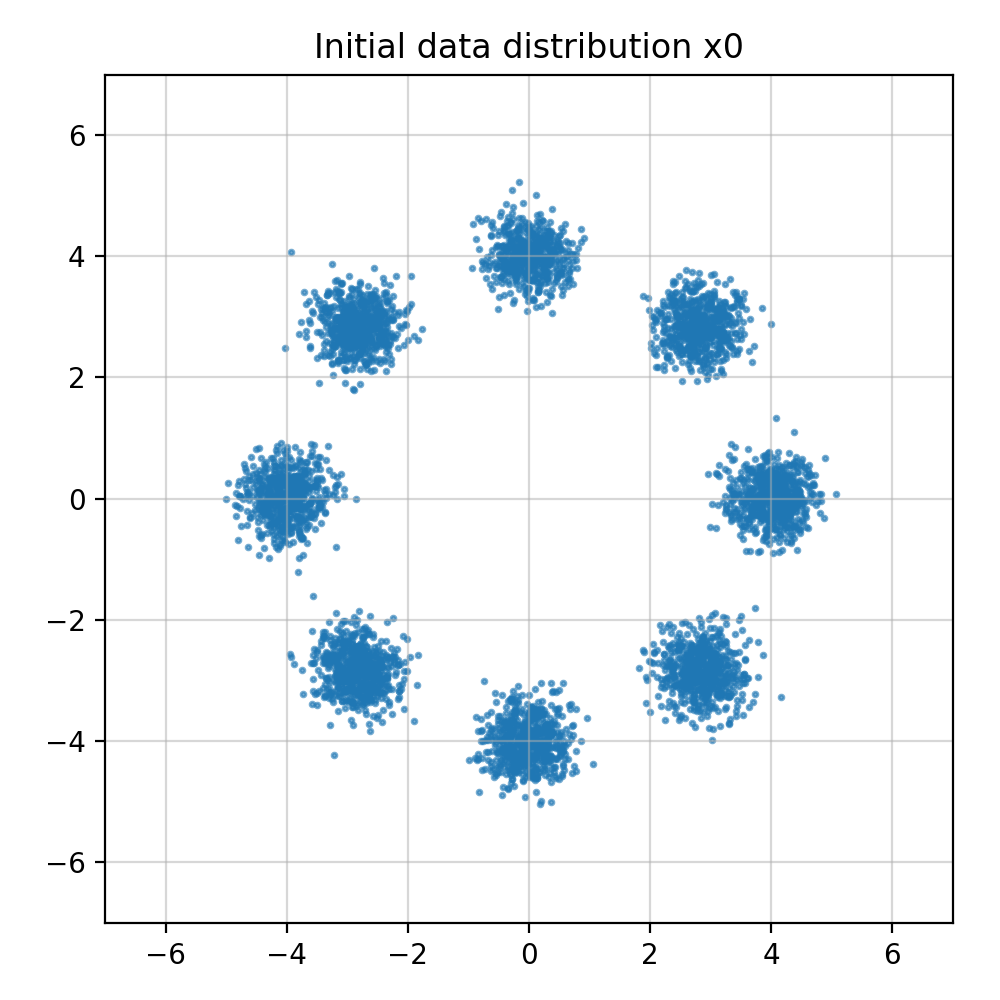}
\end{subfigure}
\hfill
\begin{subfigure}{0.45\textwidth}
\includegraphics[width=\linewidth]{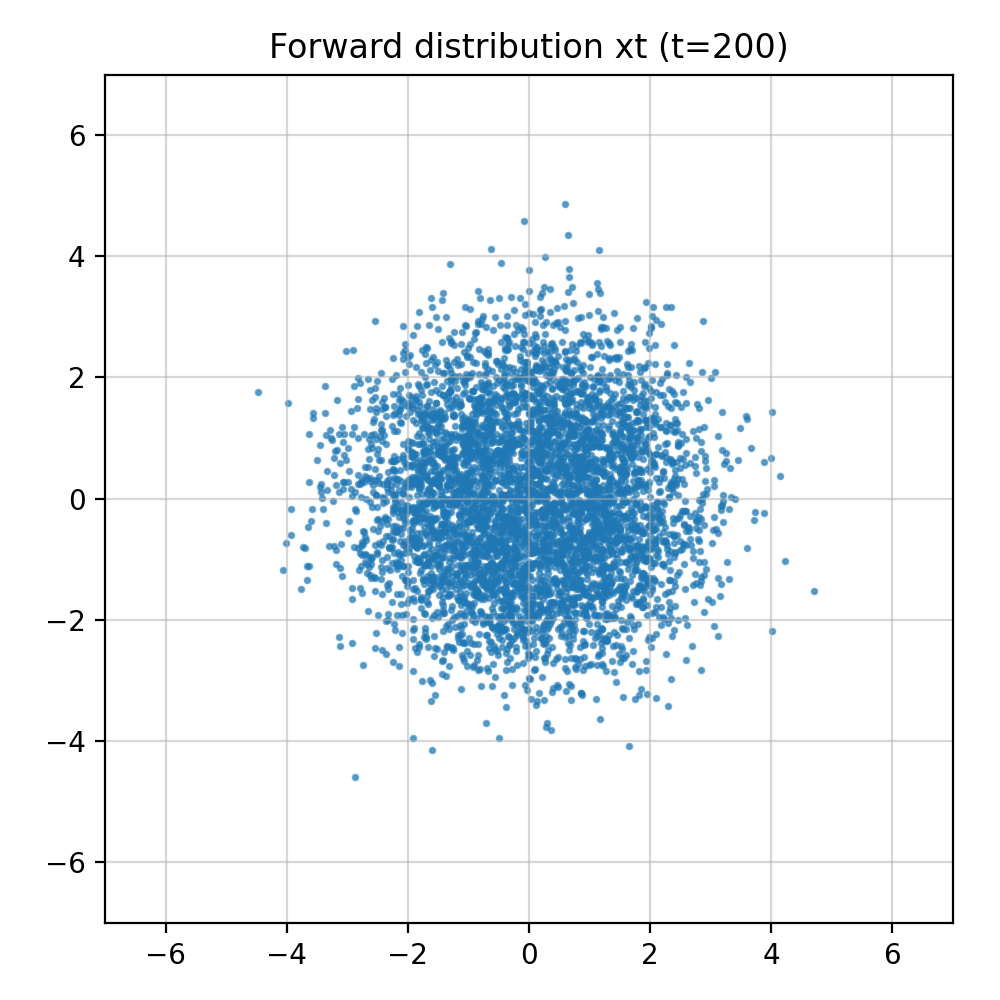}
\end{subfigure}
\caption{Toy diffusion environment based on a Ring Gaussian Mixture Model. Left shows the data distribution at $t=0$. Right shows the corresponding noisy state after forward diffusion.}
\label{fig:ring_gmm_toy}
\end{figure}

\subsection{Ring Gaussian Mixture Model}

The data distribution is a two dimensional Ring Gaussian Mixture Model. Specifically, $K$ Gaussian components are placed uniformly on a circle with radius $R$. Each component has an identical covariance $\sigma_{\text{data}}^2 I$. Each sample is generated by first drawing a class label $y \in \{1,\dots,K\}$ from a uniform distribution and then adding Gaussian noise to the corresponding component mean.

This setting has several useful properties. The conditional distribution $p(x_0 \mid y)$ and the unconditional distribution $p(x_0)$ are both available in closed form. In addition, at an arbitrary diffusion timestep $t$, the conditional and unconditional score functions $\nabla \log p_t(x_t \mid y)$ and $\nabla \log p_t(x_t)$ can be computed exactly. As a result, the optimal noise $\epsilon^{\ast}$ can be obtained analytically, which allows direct measurement of model prediction errors with respect to ground truth.

\paragraph{Implementation Details.}
The Ring GMM consists of $K=8$ components uniformly placed on a circle with radius $R=4.0$, and covariance $\sigma_{\text{data}}^2 I$ with $\sigma_{\text{data}}=0.35$. 
The diffusion process uses $T=200$ timesteps with a linear beta schedule from $\beta_1=10^{-4}$ to $\beta_T=2\times10^{-2}$. 
Models are trained with conditional dropout probability $p_{\text{drop}}=0.3$. 
The guidance scale $\omega$ is varied depending on the experiment.

\subsection{Diffusion Process and Oracle Quantities}

The diffusion process follows the standard DDPM formulation. A linear beta schedule is used and the noisy sample at timestep $t$ is generated as
\begin{equation}
x_t = \sqrt{\bar{\alpha}_t} x_0 + \sqrt{1 - \bar{\alpha}_t}\,\epsilon .
\end{equation}

Due to the structure of the Ring GMM, the conditional and unconditional score functions at each timestep $t$ are analytically tractable. A score to noise transformation is then applied to compute the optimal noise $\epsilon_t^{\ast}$. This quantity is not used during training or correction. It is used only for error decomposition and evaluation metrics in the toy experiments.

\subsection{CFG Training with Conditional Dropout}

The model is trained as a conditional diffusion model with conditional dropout for CFG training. At each training step, the conditioning label $y$ is replaced by a null token with probability $p_{\text{drop}}$ to train the unconditional prediction. The conditional and unconditional predictions share the same network and the conditioning information is provided through an embedding.

The dropout probability $p_{\text{drop}}$ controls the relative accuracy of the unconditional model. In the toy experiments, models trained with different values of $p_{\text{drop}}$ are used to analyze how the balance between conditional and unconditional prediction errors affects the behavior of CFG and CFG-OEC.

\subsection{Evaluation Protocol}

Evaluation is performed using resampled data with random seeds that are not used during training. The following error vectors are defined for analysis
\begin{equation}
\Delta_c = \hat{\epsilon}_t^{(c)} - \epsilon_t^{\ast}, \quad
\Delta_{uc} = \hat{\epsilon}_t^{(uc)} - \epsilon_t^{\ast}.
\end{equation}

The CFG sampling error is decomposed into a base term and a cross term. The magnitude and relative contribution of each term are measured quantitatively. For proxy based CFG-OEC, correction is performed using only model predictions at adjacent timesteps $t$ and $t+1$. Ground truth noise is not used for correction and is only used for evaluation.

This toy environment is designed to analyze structural misalignment between the CFG sampling rule and the training objective and to enable precise observation of how CFG-OEC mitigates this issue.

\clearpage
\section{Why Better Training Improves Sampling Despite CFG Discrepancies}

This appendix provides a theoretical explanation for an apparent paradox in classifier free guidance. Although CFG introduces a structural discrepancy between training and sampling that cannot be eliminated by optimization, improved training still leads to better sampling performance in practice. We formalize this observation by relating the training objective and the sampling error through a norm based bound.

\subsection{Training Objective and Sampling Rule}

Let $\epsilon_{t,c}^{\ast}$ and $\epsilon_{t,uc}^{\ast}$ denote the ground truth conditional and unconditional noise at timestep $t$. The model predicts conditional and unconditional noise $\epsilon_{\theta}(x_t \mid y)$ and $\epsilon_{\theta}(x_t)$ using the same network with conditional dropout.

During training, the model minimizes the expected weighted sum of squared errors
\begin{equation}
e_t^{\text{train}}
=
\mathbb{E}\Big[
p_{\text{drop}}
\|\epsilon_{\theta}(x_t) - \epsilon_t^{\ast}\|^2
+
(1 - p_{\text{drop}})
\|\epsilon_{\theta}(x_t \mid y) - \epsilon_t^{\ast}\|^2
\Big],
\label{eq:appendix_train_error}
\end{equation}
where $p_{\text{drop}} \in (0,1)$ is the conditional dropout probability used for CFG training.

At sampling time, classifier free guidance combines unconditional and conditional predictions using the guidance scale $\omega$
\begin{equation}
\tilde{\epsilon}_{\theta}(x_t \mid y)
=
(1-\omega)\,\epsilon_{\theta}(x_t)
+
\omega\,\epsilon_{\theta}(x_t \mid y).
\label{eq:appendix_cfg_rule}
\end{equation}

\subsection{Error Decomposition}

To analyze the discrepancy between training and sampling, we define the unconditional and conditional prediction error vectors
\begin{equation}
\Delta_{uc}
=
\epsilon_{\theta}(x_t) - \epsilon_t^{\ast},
\qquad
\Delta_{c}
=
\epsilon_{\theta}(x_t \mid y) - \epsilon_t^{\ast}.
\label{eq:appendix_error_vectors}
\end{equation}

Using these definitions, the squared sampling error can be written as
\begin{equation}
\begin{aligned}
e_t^{\text{sample}}
&=
\left\|
(1-\omega)\Delta_{uc}
+
\omega \Delta_{c}
\right\|^2 \\
&=
(1-\omega)^2 \|\Delta_{uc}\|^2
+
\omega^2 \|\Delta_{c}\|^2
+
2\omega(1-\omega)
\langle \Delta_{uc}, \Delta_{c} \rangle .
\end{aligned}
\label{eq:appendix_sampling_error}
\end{equation}

The first two terms correspond to the base error terms. The inner product term is the cross error. This cross error arises from the interaction between unconditional and conditional prediction errors and is not explicitly optimized during training.

\subsection{Why Training Still Improves Sampling}

Although the cross error term cannot be directly minimized by the training objective in \eqref{eq:appendix_train_error}, the overall sampling error remains controlled by the magnitude of the error vectors. This can be shown using the Cauchy–Schwarz inequality.

Let $c = 1-\omega$ and $d = \omega$ denote the sampling coefficients used in CFG. Applying Cauchy–Schwarz yields
\begin{equation}
\left\|
c\,\Delta_{uc}
+
d\,\Delta_{c}
\right\|^2
\le
(c^2 + d^2)
\left(
\|\Delta_{uc}\|^2
+
\|\Delta_{c}\|^2
\right).
\label{eq:appendix_cs_bound}
\end{equation}

The right hand side depends only on the squared norms of $\Delta_{uc}$ and $\Delta_{c}$. These quantities are precisely the components minimized by the training objective. In particular, for positive scalars $a$ and $b$,
\begin{equation}
\|\Delta_{uc}\|^2 + \|\Delta_{c}\|^2
\le
\frac{1}{\min(a^2,b^2)}
\left(
\|a\Delta_{uc}\|^2
+
\|b\Delta_{c}\|^2
\right).
\end{equation}

Choosing $a=\sqrt{p_{\text{drop}}}$ and $b=\sqrt{1-p_{\text{drop}}}$ directly relates this bound to the training error in \eqref{eq:appendix_train_error}. Substituting into \eqref{eq:appendix_cs_bound} yields
\begin{equation}
e_t^{\text{sample}}
\le
\frac{(1-\omega)^2 + \omega^2}{\min(p_{\text{drop}},1-p_{\text{drop}})}
\, e_t^{\text{train}} .
\label{eq:appendix_final_bound}
\end{equation}

This result shows that although the cross error term cannot be eliminated by training, the sampling error remains upper bounded by the training error scaled by a factor that depends on $\omega$ and $p_{\text{drop}}$. As training progresses and both $\|\Delta_{uc}\|$ and $\|\Delta_{c}\|$ decrease, the absolute magnitude of the sampling error also decreases.

\subsection{Effect of Orthogonal Error Correction}

If the unconditional and conditional error vectors are orthogonal, that is
\begin{equation}
\langle \Delta_{uc}, \Delta_{c} \rangle = 0 ,
\end{equation}
the cross error term in \eqref{eq:appendix_sampling_error} vanishes. In this case, the sampling error reduces to
\begin{equation}
e_t^{\text{sample}}
=
(1-\omega)^2 \|\Delta_{uc}\|^2
+
\omega^2 \|\Delta_{c}\|^2 ,
\end{equation}
which admits a strictly tighter bound
\begin{equation}
e_t^{\text{sample}}
\le
\frac{\max((1-\omega)^2,\omega^2)}{\min(p_{\text{drop}},1-p_{\text{drop}})}
\, e_t^{\text{train}} .
\label{eq:appendix_orthogonal_bound}
\end{equation}

CFG-OEC explicitly targets this alignment issue by enforcing approximate orthogonality between $\Delta_{uc}$ and $\Delta_{c}$ during sampling. This correction does not replace training. Instead, it complements training by tightening the sampling error bound that cannot be improved through optimization alone.

\clearpage
\section{Additional Visual Results}
\begin{figure*}[t]
    \centering
    \includegraphics[width=0.88\textwidth]{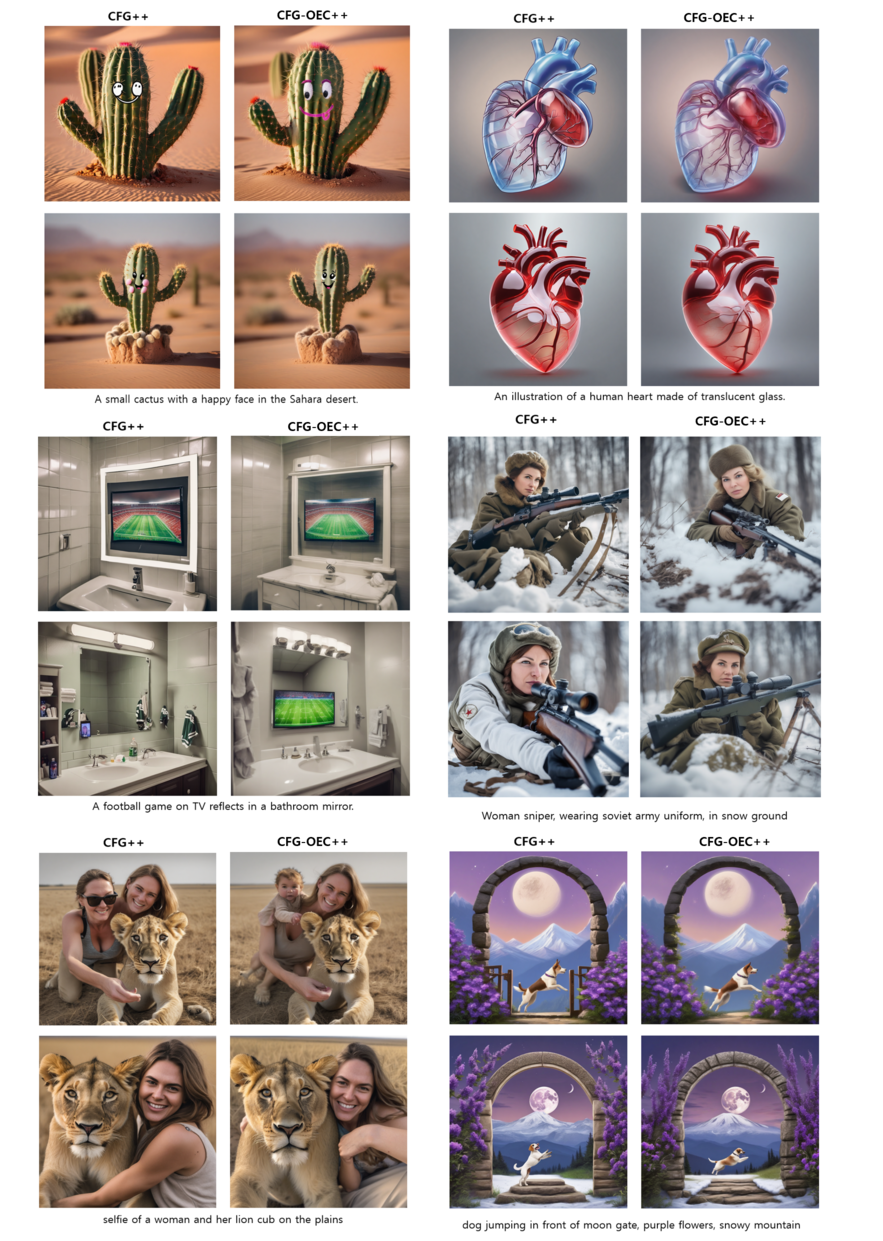}
    \caption{The corresponding SDXL ($\omega$ = 0.6) T2I images generated using CFG++ (left) and CFG-OEC++ (right) demonstrate improvements. The generated images exhibit fewer artifacts and better prompt alignment.}
    \label{fig:placeholder}
\end{figure*}

This section presents an extended visual comparison between CFG++~\cite{cfgpp} and CFG-OEC++. To verify this comparison on high-quality images, we used the latest T2I model, Stable Diffusion XL ~\cite{stablexl}, using the default DDIM sampler~\cite{ddim} with the same seed, prompt and guidance scale ($\omega$ = 0.6) 50NFEs.\\
Looking at \cref{fig:placeholder}, the cactus image (upper left) compared to existing methods, the cactus shows a more realistic representation of the face, a more distinct smile, and the artificial elements on the cheek have been removed. Furthermore, the transparency of the glass and the blood image (upper right) highlight the transparency of the glass, and the blood vessels appear more naturally connected. In the original CFG++ method, soccer matches were not generated, but our method successfully produced them. Similarly, while the original CFG++ often yielded misaligned mirrors and TV screens, our method achieved proper alignment. As illustrated in the middle-right image, conventional methods (e.g., CFG++) often exhibit artifacts where the firearm's geometry is distorted by features from the subject's clothing, resulting in an unnatural curvature. In contrast, our proposed method generates a more realistic, free from such artifacts. Furthermore, the image in the lower left demonstrates that our approach produces a more realistic scene with enhanced detail in the landscape (e.g., the plains) compared to the baseline. A critical failure mode of the existing method is observable in the lower right image, which incorrectly rendered two tails and three paws on the subject dog. However, our method correctly generates the anatomically accurate count of one tail and two visible paws.\\

\clearpage
\section{Evolution of Denoised Estimates}
\begin{figure*}[t]
    \centering
    \includegraphics[width=0.88\textwidth]{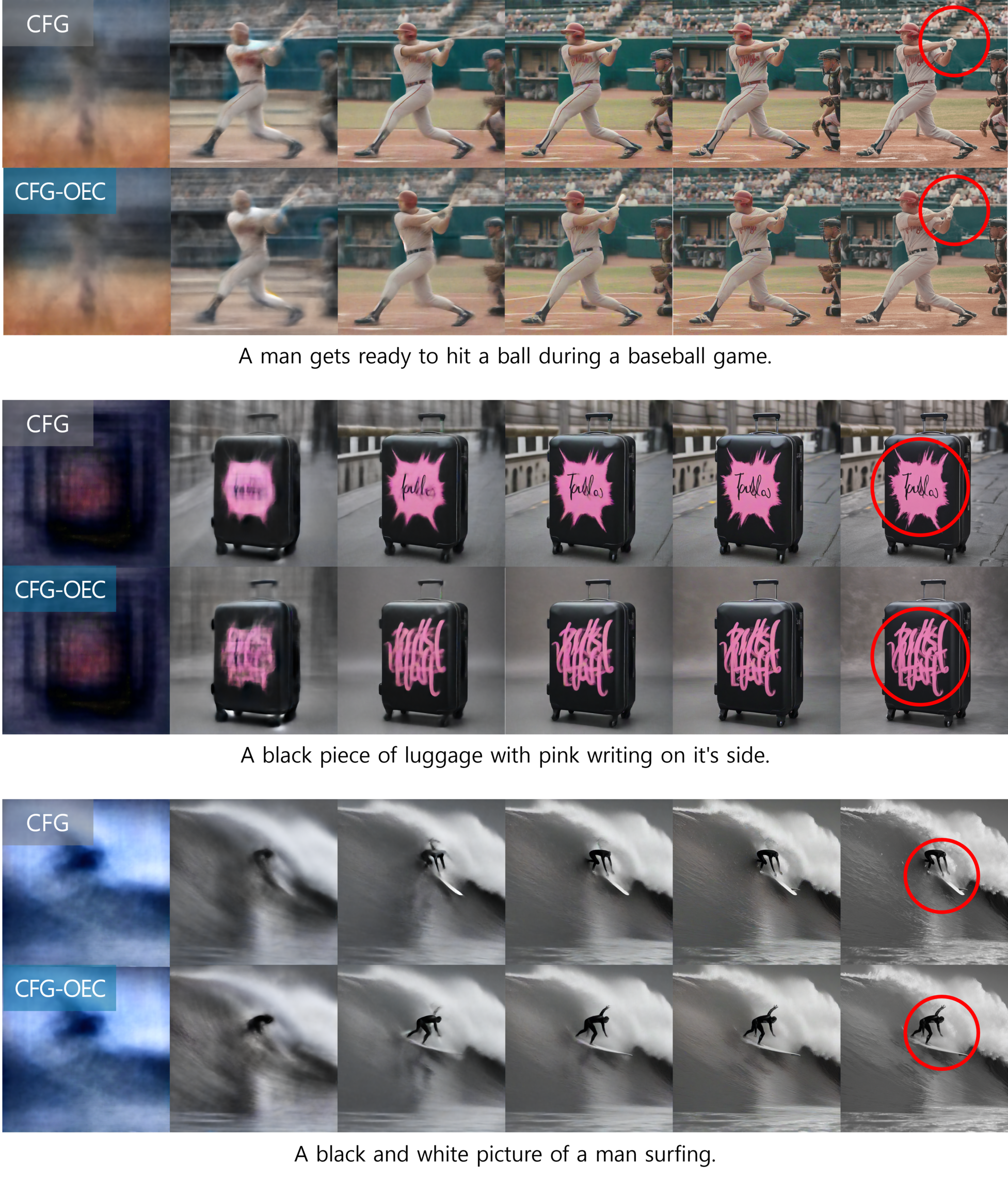}
    \caption{Evolution of denoised estimates generated by SDXL ($\omega$ = 5.0). CFG-OEC (bottom) shows better adherence to text alignment during the denoising process compared to CFG (top).}
    \label{fig:denoised}
\end{figure*}

In this section, we provide a comparison of the progression of denoised estimates generated by CFG-OEC and CFG~\cite{cfg}. We employ Stable Diffusion XL (SDXL) with the DDIM sampler using the dynamic method of CFG-OEC with 50 NFEs and guidance scale ($\omega$) to 5.0. \\
In \cref{fig:denoised}, the top image, corresponding to the prompt ``A man gets ready to hit a ball during a baseball game.'', demonstrates our method's ability to preserve key objects. While the CFG method fails to retain the baseball bat, our proposed method successfully maintains its presence throughout the sampling process. Similarly, In the middle image, for the prompt "A black piece of luggage with pink writing on it's side.'' the baseline method fails to apply the ``pink writing'' attribute, producing an incoherent pink pattern instead of text. Conversely, our method generates distinct pink lettering. Finally, the bottom image, which shows ``A black and white picture of a man surfing.'' reveals a compositional error in the CFG. The CFG method generates an incorrect configuration, with the surfer positioned away from the surfboard. Our method, in contrast, correctly places the surfer on the board, resulting in a more realistic and coherent scene.\\
Collectively, while the baseline CFG often converges to flawed representations, such as omitting objects, failing to render details, or generating incorrect spatial relationships, our method maintains a more accurate generative trajectory. The corrective mechanism of our approach appears to provide a continuous preventive mechanism that prevents the model from deviating into incorrect states where concepts are improperly merged or details are lost. 

\end{document}